\documentclass[letterpaper, 10 pt, conference]{ieeeconf}  % Comment this line out if you need a4paper

\IEEEoverridecommandlockouts                              % This command is only needed if 
\usepackage{graphics} % for pdf, bitmapped graphics files
\usepackage{epsfig} % for postscript graphics files
\usepackage{mathptmx} % assumes new font selection scheme installed
\usepackage{times} % assumes new font selection scheme installed
\usepackage{amsmath} % assumes amsmath package installed
\usepackage{amssymb}  % assumes amsmath package installed

\usepackage{multirow}
\usepackage{booktabs}
\usepackage{verbatim}
\usepackage[T1]{fontenc}
\usepackage{verbatim}
\usepackage{caption}
\usepackage[table]{xcolor}

\title{\LARGE \bf
Audio-Visual Grounding Referring Expression for Robotic Manipulation
}

\author{   {Yefei Wang$^*$, Kaili Wang$^*$, Yi Wang$^*$, Di Guo, Huaping Liu, Fuchun Sun}% <-this % stops a space
\thanks{$^*$ denotes the equal contributions. All of the authors are with the Department of Computer Science and Technology, Tsinghua University, Beijing, China. e-mail: hpliu@tsinghua.edu.cn}% <-this % stops a space
}

\begin{document}

\maketitle
\thispagestyle{empty}
\pagestyle{empty}

%%%%%%%%%%%%%%%%%%%%%%%%%%%%%%%%%%%%%%%%%%%%%%%%%%%%%%%%%%%%%%%%%%%%%%%%%%%%%%%%
\begin{abstract}

% \textcolor{blue}{In this paper, we present an audio-visual robot system that can interact and recognize with target objects in the scene through natural language instructions. Specifically, we build a referring expression model and an audio recognition model. The former distinguish the positional relationship of objects in the scene, while the latter assist the robot detect with the help of audio information when the visual information is not enough to complete the task. Both the audio and visual information are leveraged to interpret the referring expression in robotic manipulation.Finally, we also suggest a multi-modal dataset to verify the effectiveness of our entire system.}

Referring expressions are commonly used when referring to a specific target in people's daily dialogue. In this paper, we develop a novel task of audio-visual grounding referring expression for robotic manipulation. The robot leverages both the audio and visual information to understand the referring expression in the given manipulation instruction and the corresponding manipulations are implemented. To solve the proposed task, an audio-visual framework is proposed for visual localization and sound recognition. We have also established a dataset which contains visual data, auditory data and manipulation instructions for evaluation. Finally, extensive experiments are conducted both offline and online to verify the effectiveness of the proposed audio-visual framework. And it is demonstrated that the robot performs better with the audio-visual data than with only the visual data. 

\end{abstract}

%%%%%%%%%%%%%%%%%%%%%%%%%%%%%%%%%%%%%%%%%%%%%%%%%%%%%%%%%%%%%%%%%%%%%%%%%%%%%%%%
\section{INTRODUCTION}

Referring expressions are commonly used when people talking with each other specifying some particular objects in the scene. For example, "the cup next to the computer", "the brown bag on the chair", etc. By understanding the referring expression, the target object can be localized in the scene given natural language description. Different from traditional visual perception tasks which have predefined object labels, the referring expression task is faced with more complex language and visual semantics making it a more challenging task. Currently, the referring expression task has aroused the attention from both the computer vision and natural language communities. And various methods and datasets for referring expression tasks are proposed \cite{qiao2020referring}\cite{yu2016modeling}\cite{kazemzadeh2014referitgame}\cite{hu2017modeling}.

However, existing referring expression tasks are mostly subjected to static scenarios and the referred objects have to be within the image. Considering the navigation ability of the robot, it will be of more practical usage if the referring expression task can be implemented while the robot exploring the environment. Recently, Ref. \cite{qi2020reverie} introduces a new task named \textit{Remote Embodied Visual referring Expression in Real Indoor Environments (REVERIE)}, in which the target object might not in sight at first and the agent needs to explore the environment to identify the target object given natural language instruction. It is a more challenging task as it additionally introduces the action into referring expression task and requires the joint learning of vision, language and action together. Thereafter, to achieve the REVERIE task, Ref. \cite{gao2021room} exploits the linguistic and visual clues together with commonsense knowledge to generate the exploration action for the agent.

Although the navigation ability of the agent is considered in REVERIE, the manipulation ability, one of the most important charateristics of robot, is ignored. Ref.\cite{misra2016tell} makes some early attempts on this topic, and Ref.\cite{paul2018temporal} develops temporal grounding graphs for language understanding with manipulators. To solve the referring expression problem, a robotic system INGRESS \cite{shridhar2020ingress} is proposed, in which the robot is able to pick and place objects following natural language instructions. And a POMDP model is used to eliminate ambiguity and facilitate to ground natural language referring expressions in the robotic manipulation scenario. Furthermore, Ref. \cite{zhang2021invigorate} introduces a robot system INVIGORATE which enables the robot to grasp a target object from the clutter where occlusions might exist given human instructions. Both the model-based POMDP planning and data-driven deep learning methods are leveraged for the robot to continuously interact with the environment and human. Ref. \cite{ahn2020visually} additionally takes the manipulation history into consideration when visually grounding a series of text instructions for robotic manipulations.
\begin{figure}[t] 
	\centering 
	\includegraphics[width=0.9\linewidth]{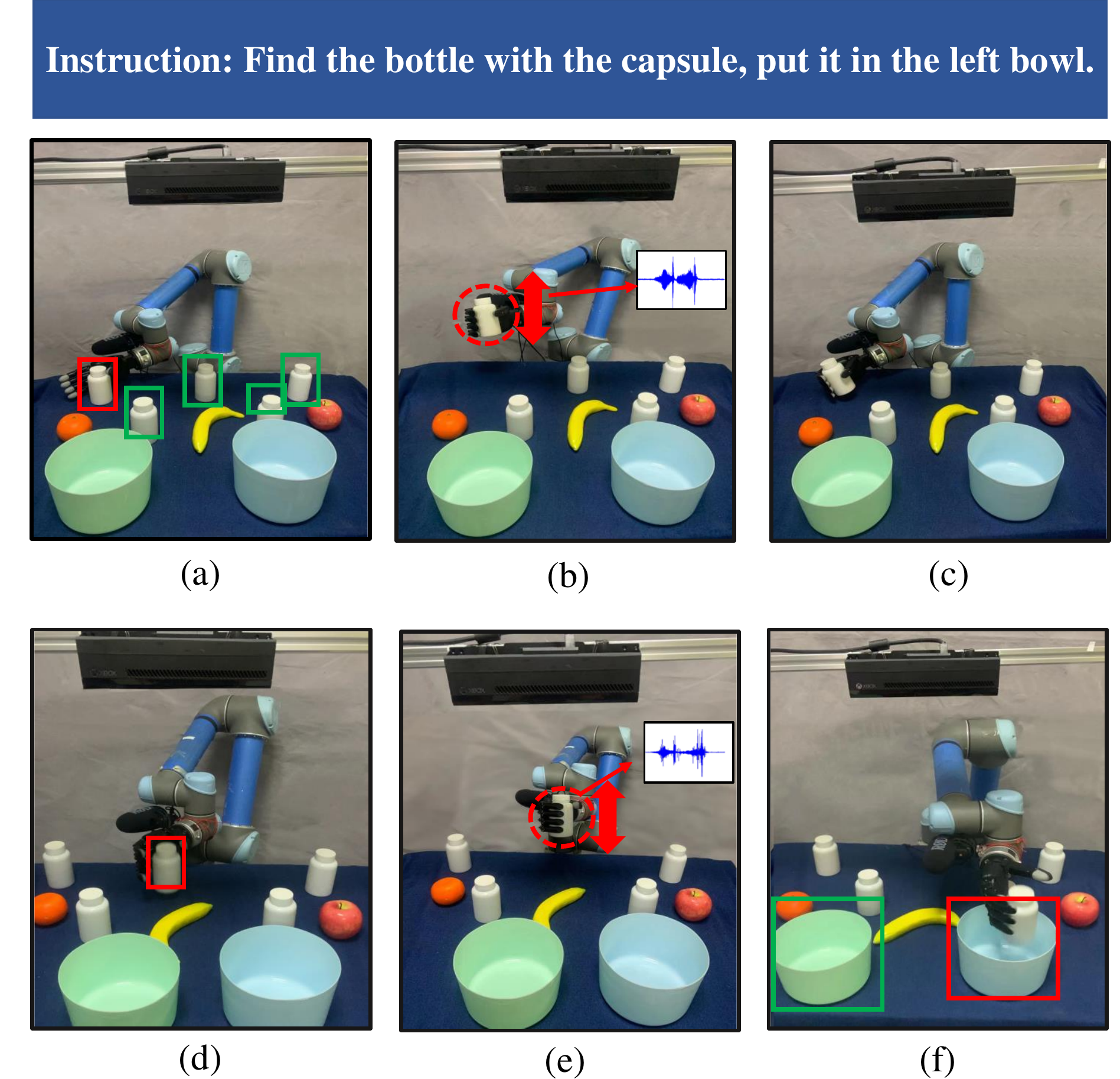} 
	\caption{An illustration of the proposed Audio-Visual Grounding Referring Expression for Robotic Manipulation. In this example, the robot receives a natural language instruction and manipulate with the objects accordingly. (a) Start to pick up the bottle. (b) Perform actions to collect audio information for judgment. (c) If it is not the target object, put the bottle back. (d) Pick up the second bottle in the middle. (e) Perform actions to collect audio information for judgment. (f)If it is the target object, put it in the target bowl.}
	\label{Audio-visual Task}
\end{figure}

All the above mentioned tasks are only based on visual information. However, in the real-world environment, other sensory modalities can also provide complementary perception information \cite{8793860}\cite{chen2021semantic}\cite{magassouba2018multimodal}\cite{tatiya2019}, among which sound is widely used in many tasks. In \cite{strahl2018hear}\cite{jin2019open}, the sound of the object is used to distinguish identical bottles which have different contents in them. Besides sound information, Ref. \cite{jonetzko2020multimodal} also introduces the tactile information to improve the auditory classification performance. In \cite{inceoglu2018failure}, the sound information can work together with visual information to detect failure in robotic manipulation. Additionally, a type of audio-visual embodied navigation task is proposed recently, in which the agent navigates to a sounding object by leveraging both visual and auditory data \cite{gan2020look}\cite{chen2020soundspaces}.

In this paper, we propose a new robotic manipulation task for audio-visual embodied referring expression (Fig. \ref{Audio-visual Task}). Both the audio and visual information are leveraged to interpret the referring expression in robotic manipulation. For example, "find the bottle with the capsule and put it in the left bowl", which requires the robot to firstly localize all the bottles on the table and then identify whether there is capsule in the bottle by listening to the sound when manipulating the bottle. After finding the target bottle, the bottle is put in a target place. The main contributions of the paper are summarized as the following:
\begin{itemize}
	\item We develop a novel task of audio-visual grounding referring expression for robotic manipulation, where both audio and visual information are used for interpret the referring expressions.
	\item We propose an audio-visual framework which can be used for both visual localization and audio recognition for the robot to implement manipulation instructions.  
	\item We collect a multi-modal dataset and conduct extensive experiments to verify the effectiveness of the proposed framework both offline and online.
	
\end{itemize}

The remainder of the paper is organized as follows. In Section \ref{PROBLEM FORMULATION}, the problem formulation is described. And in Section \ref{ARCHITECTURE}, we present an overview of the task architecture. Section \ref{METHOD} describes the implementation details of the proposed method. Section \ref{DATASET} presents the establishment of the dataset. And then, the experimental results are analyzed in Section \ref{EXPERIMENTAL RESULTS}. Finally, we conclude the paper in Section \ref{CONCLUSION}.

\section{PROBLEM FORMULATION}
\label{PROBLEM FORMULATION}
The goal of the proposed task is to enable the robot to accomplish manipulation tasks following complex natural language instructions, and both the audio and visual information are leveraged to interpret the referring expression. 

Concretely speaking, we denote the given natural language instruction as $\mathcal{I} = \{w_1, w_2, ... ,w_T\}$, where $w_i$ denotes the $i$-th word in the instruction. The agent is expected to follow the instruction to manipulate properly. Specifically, both the visual and audio information are required to fully interpret the referring expression in the instruction. For example, in the referring expression "a bottle with pills in it", the robot firstly localizes the bottles in the scene by understanding the captured visual information, while it is impossible to know whether there is pills in the bottle. And then the robot could shake the bottle and identify the target bottle by analyzing the sound of shaking. With correct comprehension of the referring expression in the given instruction, the robot is able to implement the manipulations accordingly. 

\begin{comment}

In this task, we designed the R-model and auditory recognition model S-model for referential expression, and integrated them into our robotic arm manipulation model M-model. In the instruction expression model, R-model can parse out the target object and the corresponding image coordinates according to the manipulation instructions and task scene. Our M-model can select the action according to the target coordinates, accuse the UR arm, and take the bottle in the scene. . And according to the needs of object discrimination, select different collection actions to collect audio information, call S-model for auditory recognition, find the correct object, and improve the task goal. The main challenge of the multi-modal task we designed is whether the robot can effectively identify objects based on language manipulation instructions, combined with visual information and auditory information, and continuously generate action sequences to complete the target task.
??...
\end{comment}
\section{ARCHITECTURE}
\label{ARCHITECTURE}

\begin{figure}[t] 
	\centering 
	\includegraphics[width=0.9\linewidth]{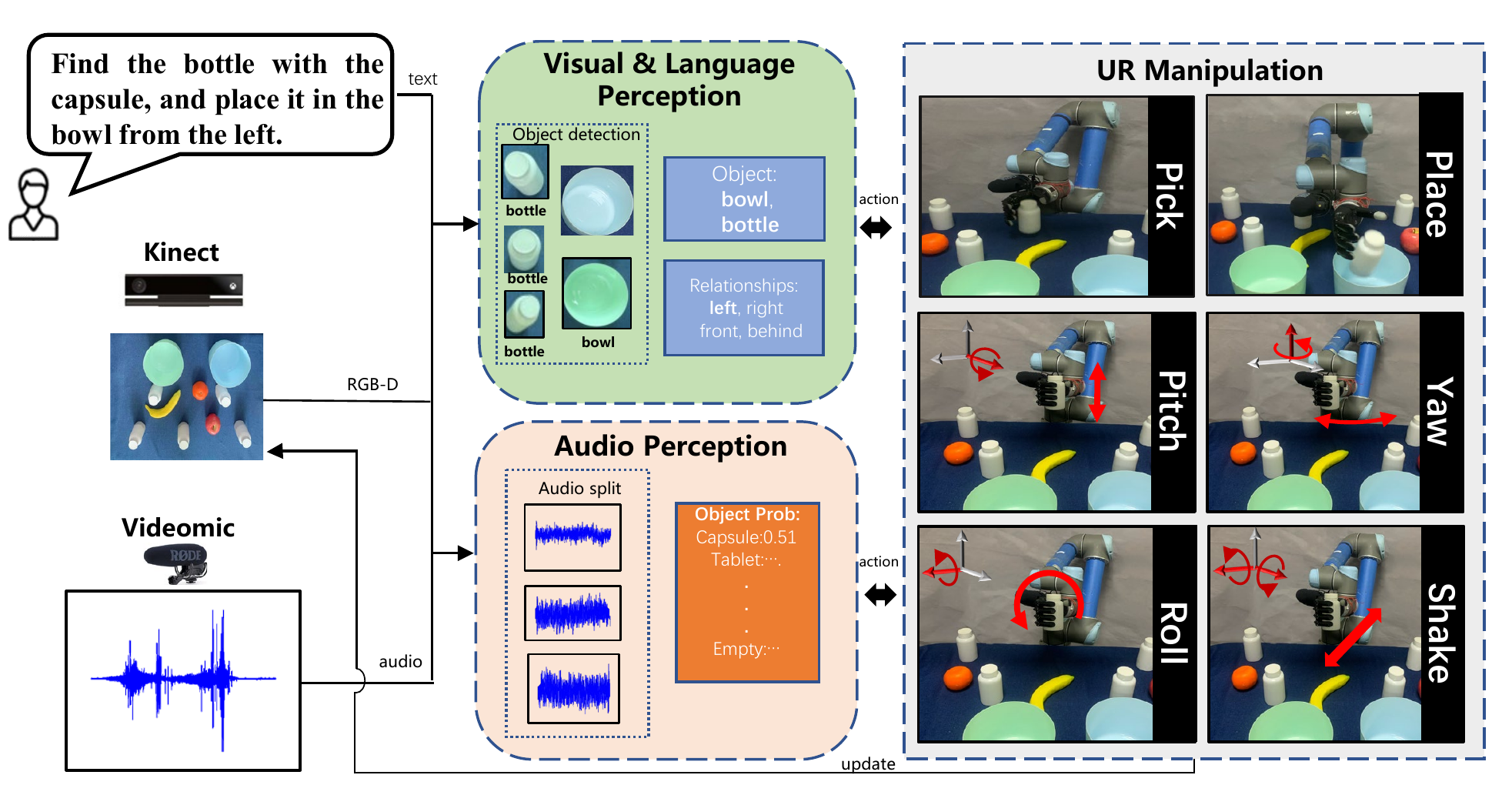} 
	\caption{System Block Diagram. The entire system relies on language, vision, and audio as input, combined with visual-language and audio recognition modules for processing, and uses different manipulation actions to interact with the environment.}
	\label{System Block Diagram}
\end{figure}

The architecture of the proposed system is demonstrated in Fig. \ref{System Block Diagram}, which is composed of a visual-language perception module, audio perception module and manipulation module. At first, the visual information together with the textual instruction is fed into the visual-language module to localize possible targets of the referred objects. As it is difficult to identify the target object with only visual information, the manipulation module will then be activated to implement the different actions to generate sound information. Meanwhile, the sound can be recorded by the equipped auditory sensor. With the collected audio information, the audio recognition module can analyze the audio to recognize the target object that is referred in the instruction. Finally, the robot is able to correctly ground the textual instruction into the manipulation scenario using both the audio and visual information, and the manipulation module will execute proper actions accordingly. 

\section{METHOD}
\label{METHOD}
% \textcolor{blue}{As shown in Fig. \ref{4}, including visual understanding part, language understanding part, auditory processing part and operation part \cite{hu2017modeling} \cite{kazemzadeh2014referitgame} \cite{liu2017referring} \cite{magassouba2018multimodal} \cite{tatiya2019} \cite{8461187}. Specifically, a visual encoder and a language encoder are used to extract the multi-scale features of the input image and text, and then they are combined to obtain the position of the target object with visual semantics. Then the robot arm selects the action according to the target position, controls the UR robot arm, and grabs the bottle in the scene. And according to the discriminating needs of the object, select different collection of mobile phone audio information, combined with the auditory processing module  \cite{tatiya2019} \cite{lakomkin2019}, find the correct object, and complete the target task.}

\begin{figure*}[h] %%???????????????????????bp????
	\centering  %?????????
	\includegraphics[width=0.9\linewidth]{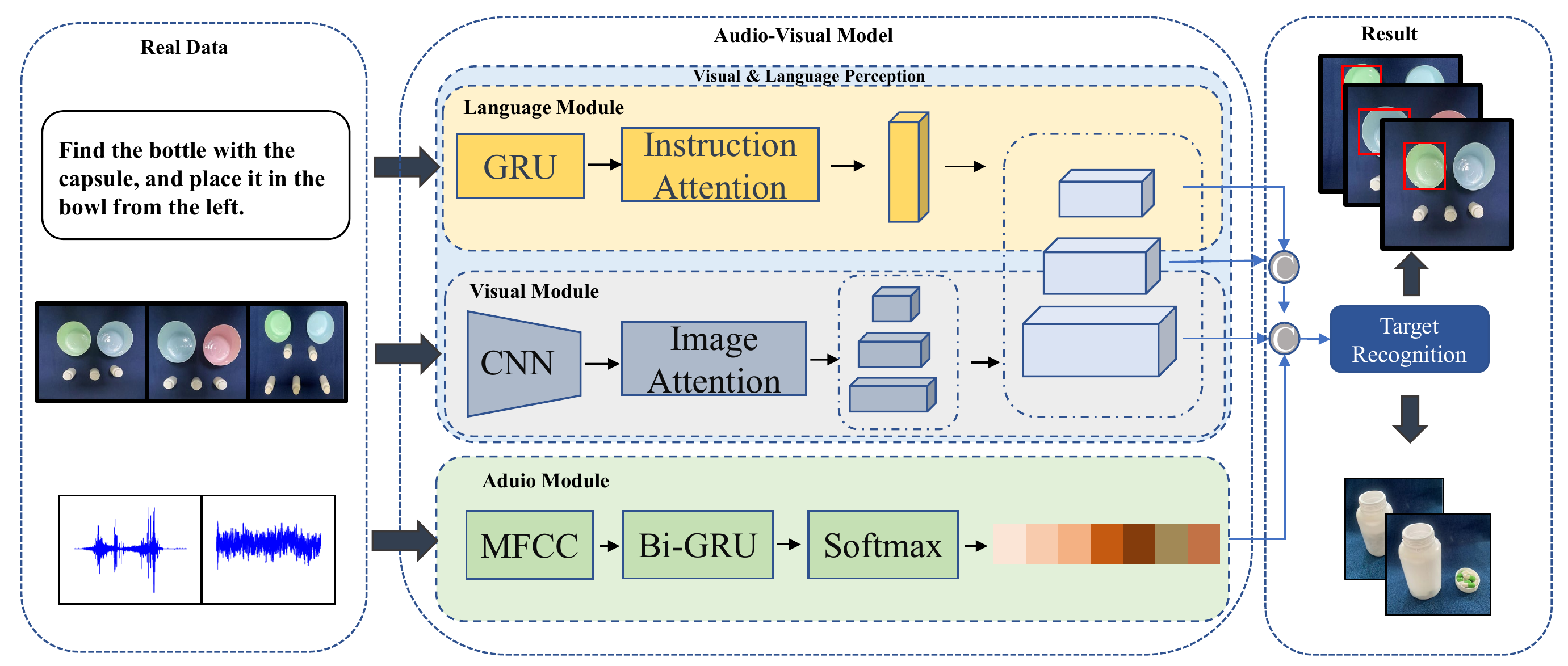} %%???????
	\caption{The structure of the proposed model for audio-visual model. This model consists of three main modules: the language module, the visual module and the auditory module.}
	\label{4}
\end{figure*}

As shown in Fig. \ref{4}, the proposed framework is composed of a visual-language module, an auditory module and a manipulation module. To understand the referring expression in the given instruction, the visual-language module is firstly implemented to localize possible target objects, and the audio module is used to further recognize the target object. The manipulation module is used to throughout the process to achieve the manipulation instruction. 

\subsection{Language model}
Inspired by \cite{yu2018mattnet}, we use a Bi-GRU to encode each word in the natural language instruction $\mathcal{I}=\{w_1, w_2,...,w_T\}$. Specifically, the one-hot embedding is firstly used to embed each word, and then a Bi-GRU is adopted to further encode the text, where the concatenation of the hidden vectors from both directions represents each word. 
\begin{equation}
e_{t}=\text { One-hot }\left(w_{t}\right)
\end{equation}
\begin{equation}
\overrightarrow{h}_{t}=\operatorname{GRU}\left({e}_{t}, \overrightarrow{h}_{t-1}\right)
\end{equation}
\begin{equation}
\overleftarrow{h}_{t}=\operatorname{GRU}\left({e}_{t}, \overleftarrow{h}_{t+1}\right) 
\end{equation}
\begin{equation}
h_{t}=\left[\overrightarrow{h}_{t}, \overleftarrow{h}_{t}\right]
\end{equation}

Instead of regarding the instruction as a whole, we try to decompose the textual instruction into subject appearance, location and relationship information for comprehension. Therefore, we define the trainable vectors $s_m$, where $m \in \{subj, loc, rel\}$, to represent the attention on each word from each perspective and the feature vector for the entire instruction can be obtained as a weighted sum of each word's embedding. 

\begin{equation}
a_{m, t}=\frac{\exp \left(s_{m} h_{t}\right)}{\sum_{k=1}^{T} \exp \left(s_{m} h_{k}\right)} 
\end{equation}
\begin{equation}
q_{m}=\sum_{t=1}^{T} a_{m, t} {e}_{t}
\end{equation}
And the concatenation of the instruction features from each perspective of $m \in \{subj, loc, rel\}$ forms the textual instruction feature $f_t=[q_{subj}, q_{loc}, q_{rel}]$.

\subsection{Visual Model}
In terms of visual features, we use Darknet-53 and feature pyramid network to extract the hierarchical visual features $f_{v_{i}}, i=1,2,3$ of the input image.  $f_{v_{i}}$. The size of the input image is $256 \times 256$, and the resolution of the three spatial feature vectors is $8 \times 8 \times 1024, 16 \times 16 \times 512$, and $32 \times 32 \times 256$ respectively. 

For each level, the visual-language feature can be fused as: 
\begin{equation}
f_{m_i}=\sigma\left(f_{t} \mathbf{W}_{\mathrm{t}}\right) \odot \sigma\left(f_{v_{i}} \mathbf{W}_{v_{i}}\right)
\end{equation}
where $\mathbf{W}_{\mathrm{t}}$ and $\mathbf{W}_{v_{i}}$ are projection weight matrices, $\sigma$ is the Leaky RelU function and $\odot$ denotes the dot-multiplication. For features of different level, we use a $2\times2$ upsampling operation to map them to the same dimension \cite{luo2020multi}.

With the visual-language fused feature, we would like to localize the target object in the scene for the manipulation. Firstly, we use the Yolo network to detect possible objects in the frame and generate the visual feature $r_{loc}$. And then the visual-language feature from different levels are concatenated to the visual-language feature $f_m$ and visual feature $r_{loc}$ are used to calculate the attention weight for each area. The area with the largest score is the most suitable location for the manipulation.

\begin{equation}
t=\varphi\left(\left(W_{v} f_{m}+b_{v}\right) \otimes\left(W_{t} r_{l o c}+b_{t}\right)\right) 
\end{equation}
\begin{equation}
\beta=\operatorname{softmax}(t)
\end{equation}
\begin{equation}
\mathrm{u}_{l o c}=\beta \otimes f_{m}
\end{equation}
\begin{equation}
s_{l o c}=\mathcal{D}\left(\mathrm{u}_{l o c}, \mathrm{r}_{l o c}\right)
\end{equation}

We use the center point as predicted location for the target object and the image coordinates are converted into the robotic coordinates for the manipulation.

\subsection{Auditory Model}
We recognize the type of the target object by analyzing the sound generated when the robot is manipulating the object. The MFCC features \cite{8794166} are selected to represent the collected sound for it shows robustness to suppress the noise generated during the manipulation. To obtain the MFCC feature, we use the Hamming window with a window size of 30ms and a step size of 15ms. Eventually, 21 Mel coefficients corresponding to the spectrum are obtained. And then the extracted MFCC features are fed into a Bi-GRU network following a Softmax classifier to recognize the material. 

In real experimental environment, there is a lot of noise generated when the robot is running, so a noise suppression algorithm is designed to filter the collected audio data. We set a threshold that is dertermined by some common signal envelope method to remove the signal that is under the threshold.

% As is shown in Fig. \ref{Audio area selection}, the areas where the audio signal does not exceed the minimum signal strength are marked as the yellow area, and are removed. Only the signals in green area are used for training. We use the common signal envelope method \cite{sharma2020trends} \cite{kauppinen2002audio} to find the appropriate threshold to remove redundant sound areas.

% \begin{figure}[h] %%???????????????????????bp????
% 	\centering  %?????????
% 	\includegraphics[width=0.9\linewidth]{figure/auidoplot.pdf} %%???????
% 	\caption{Audio area selection}
% 	\label{Audio area selection}
% \end{figure}
\begin{table*}[h]
	\large
	\centering
	\caption{Instruction type setting}
	\label{Problem type setting}
	\resizebox{\textwidth}{7mm}{
		\begin{tabular}{cccc}
			\hline
			%\toprule[2pt]
			Instruction type &Instruction template&Example sentence&Dataset size\\
			\hline
			Existence Instruction&Find the bottle with the
			<obj>, put it in the <relation>/<color> bowl.&Find the bottle with the
			hawthorn, put it in the left/green bowl.&288\\
			Classification Instruction&Find all the <obj>, and put it
			in the <relation>/<color> bowl .&Find all the hawthorns and
			put them in the left/green bowl.&288\\
			Exploratory Instruction&Check the bottle <relation>
			the <obj1> for <obj2>.&Check the bottle on the
			banana for hawthorn.&36\\
			\hline
			%& &hawthorn&pill&cassia&oyster\\
			% \rowcolor{green!10}existence Instruction&Find the bottle with the
			% <comparison> <obj>,put it in
			% the bowl on the <relation>/<color>. &Find the bottle with the
			% most hawthorn. put it in the
			% bowl on the left/green.&576\\
			% \rowcolor{green!10}Classification Instruction&Find all the <property>, and put
			% it in the <color> 
			% bowl on the <relation> &Find all the Bulk objects and
			% put it in the green bowl
			% on the left.&24\\
			% \rowcolor{green!10}Exploratory Instruction&Pick up the bottle on the far
			% <relation> of the <obj> and
			% check if it's a <obj>.&Pick up the bottle on the far
			% left of the book and check if
			% it's a hawthorn&96\\
			% \hline
			
	\end{tabular}}
\end{table*}
\subsection{Manipulate Model}
With the manipulation model, the robot is able to manipulate with the target object in the scene. The action space is defined as $\mathrm{A}=\left\{a_{p i c k}, a_{\mathrm{yaw}},  a_{\mathrm{roll}},  a_{\mathrm{pitch}}, a_{\text {shake }},  a_{\mathrm{place}}\right\}$, among which $a_{\mathrm{yaw}},  a_{\mathrm{roll}},  a_{\mathrm{pitch}}, a_{\text {shake }}$ are four actions that the robot can use to collect the sound of the object. 

When the system detects a target object referred in the instruction, the $a_{\mathrm{pick}}$ action is selected. And then the five-finger robotic hand picks up the target object, and implement different actions to generate the sound of the object. After the object is recognized, the $a_{\mathrm{place}}$ action is executed to place the object to the expected location. 

To collect the sound of the object, the robot performs the four actions $a_{\mathrm{yaw}},  a_{\mathrm{roll}},  a_{\mathrm{pitch}}, a_{\text {shake }}$ in sequence. And the sound clips corresponding to each action are fed into the sound model for recognition. The predicted class with the largest number of occurrence is taken as the class for the object.

\section{DATASET}
\label{DATASET}
\subsection{Hardware System}
\begin{figure}[h] %%???????????????????????bp????
	\centering  %?????????
	\includegraphics[width=1\linewidth]{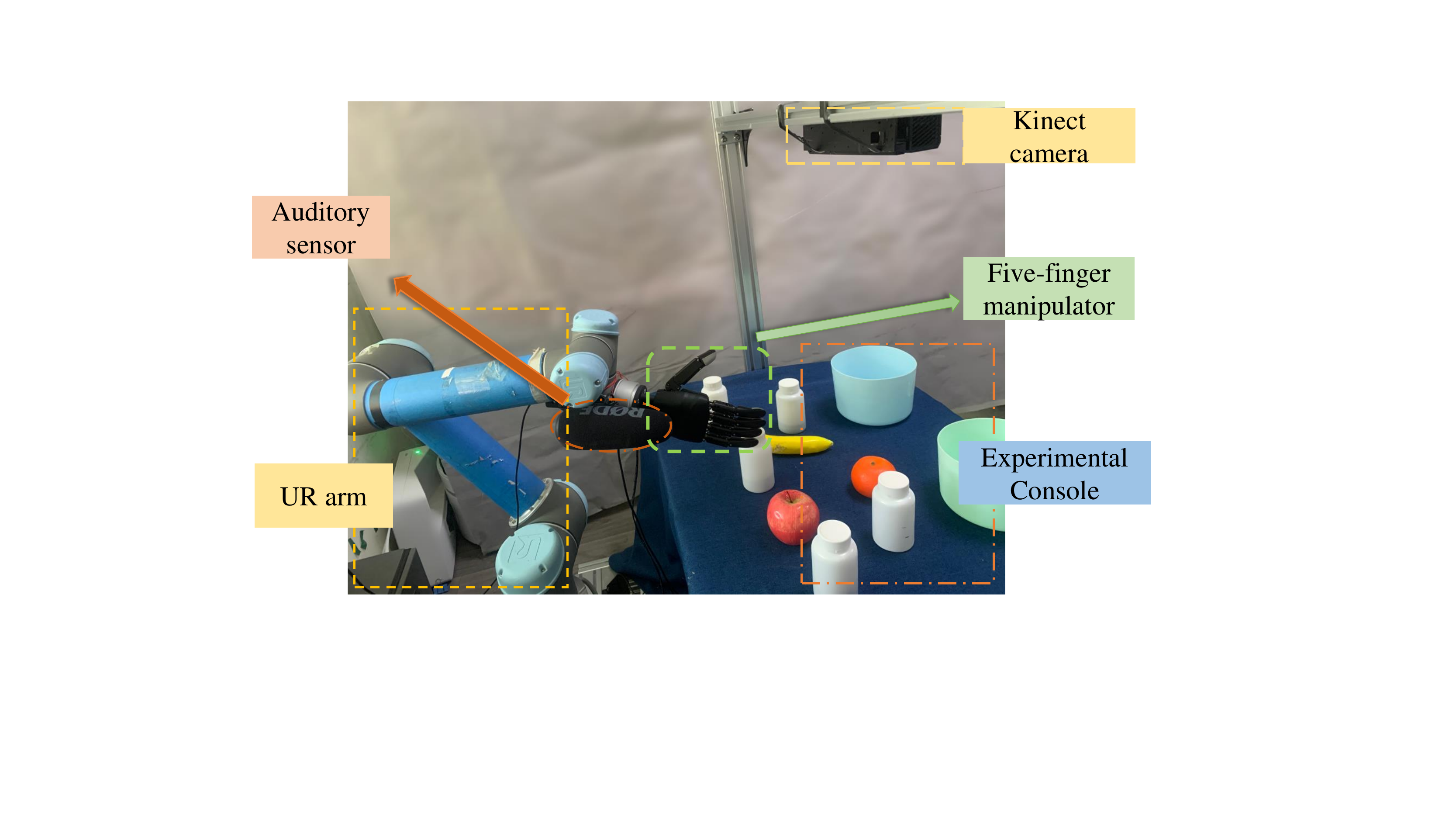} %%???????
	\caption{Hardware architecture. The hardware devices mainly include UR arms, five-finger manipulator, auditory sensor, Kinect camera, and related containers and target objects.}
	\label{Hardware architecture}
\end{figure}

The hardware system to collect the dataset is composed of a UR5 robotic arm, sound sensor, Kinect camera, and a five-finger robotic hand, running under the ROS environment with the NVIDIA 2070 GPU. An overview of the system is shown in Fig. \ref{Hardware architecture}.  

The UR5 robotic arm is a six-degree-of-freedom tandem manipulator which is used to generate different actions to interact with the bottles. A self-developed five-finger robotic hand is attached to the end of the robotic arm which is flexible to execute required manipulation actions. In this work, the robotic hand is used to grasp the bottle in the scenario. The Kinect camera is placed on the top of the table and is responsible for capturing RGB-D images for the scene. And a auditory sensor is equipped near the robotic hand. It is used to receive and record the sound of the objects when they are manipulated by the robot. 

\subsection{Auditory Dataset}
\begin{figure}[h] %%???????????????????????bp????
  \centering  %?????????
  \includegraphics[width=0.9\linewidth]{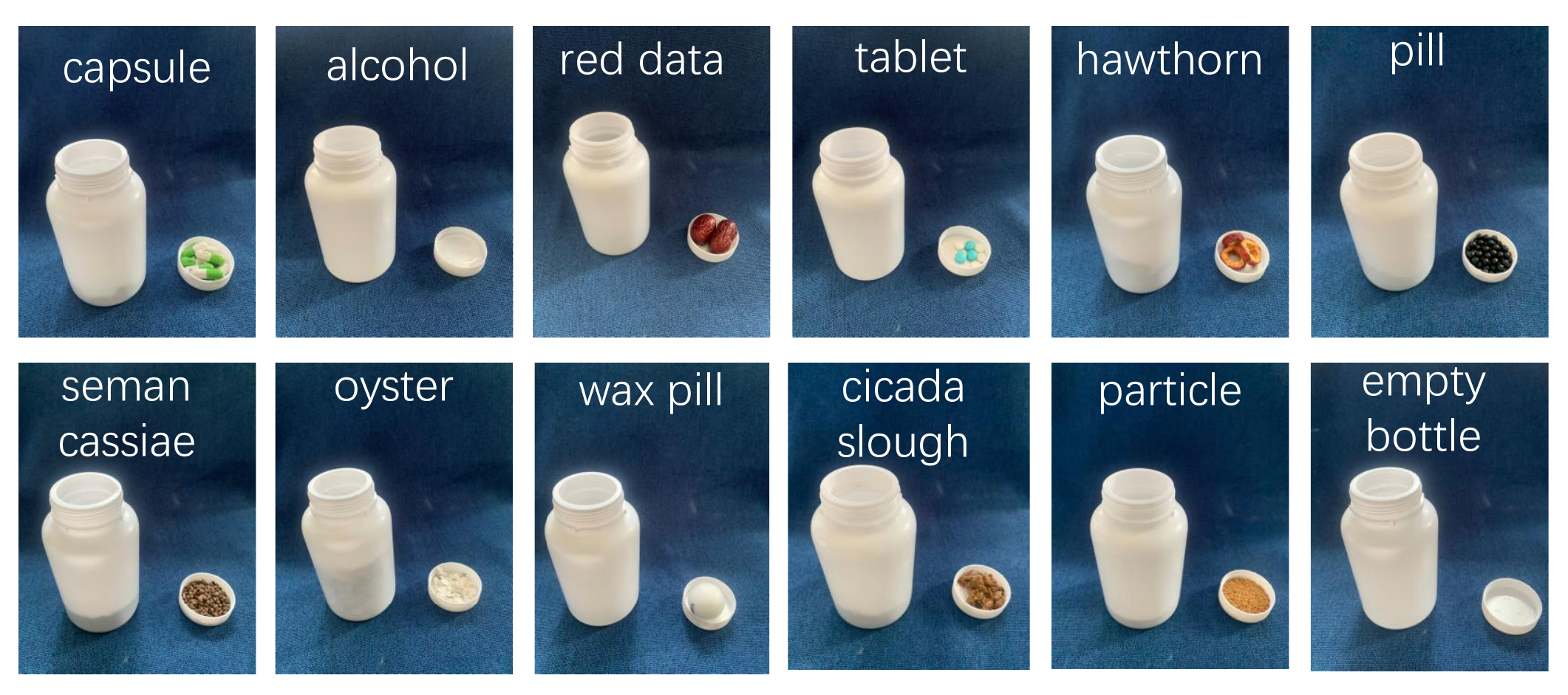} %%???????
  \caption{Object dataset. There are multiple types of object including both liquid and solid to ensure the diversity of sound data.}
  \label{Object dataset}
\end{figure}

As is shown in Fig. \ref{Object dataset}, we consider 12 types of common objects that are usually stored in the bottle. As the bottles are identical in appearance, it is difficult to distinguish different objects in bottles according to the visual information alone. Therefore, we design various actions for the robot to interact with the bottles to generate different sounds for object recognition. 

% In the entire experimental setup, we set up 12 types of common objects. Figure \ref{Object dataset} shows the container contents. All containers have the same shape, size and color, and it is difficult to distinguish categories based on visual information alone. According to human common sense actions, it is necessary to unscrew the bottle cap to view or shake it by the ear to distinguish the object category. In our experimental environment , We collect auditory information to distinguish the types of objects.

\begin{figure}[h] %%???????????????????????bp????
  \centering  %?????????
  \includegraphics[width=0.8\linewidth]{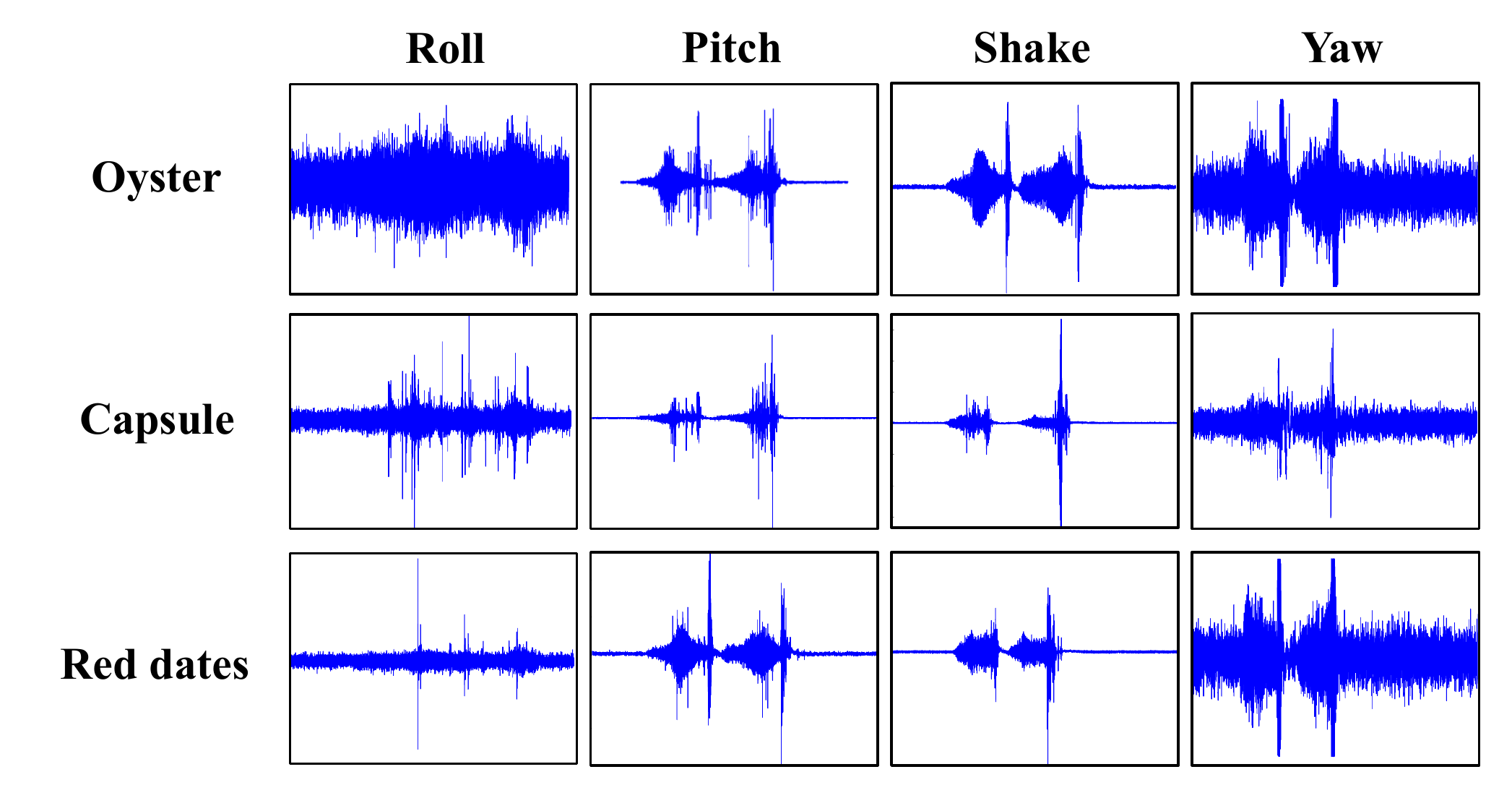} %%???????
  \caption{Sound waves analysis.}
  \label{Waveform difference analysis}
\end{figure}

Four actions, namely yaw, roll, pitch, and shake, are designed for the robot to interact with the bottle. Firstly, the bottle is manually handed over to the robotic hand and the bottle is grasped with the thumb and four fingers opposed to each other. And then the bottle is manipulated by the robotic hand. At the same time, the sound of the object will be recorded. It is noted that when implementing the yaw, roll, pitch and shake actions, the angular velocity of the hand is set to be 3.14 $rad/s$. For 12 different objects, they have been manipulated by the four actions respectively, and each with 20 times. In Fig. \ref{Waveform difference analysis}, we have visualized the sound waves of some typical objects under different actions. It can be seen that different objects have different sound characteristics and the same object could generate different sounds under different actions.

% There are ordinary environmental noises in our laboratory, such as fans, motors, and gears of robotic arms that generate most of the noise. Before shaking, an experimenter handed the drug container over and grasped it firmly with his thumb and four other fingers. In each acquisition, the force applied to the object and the finger posture remain unchanged to generate uniform data. In order to reduce the error of auditory classification, we use the joint angle of the rotating end to achieve different actions, which are defined as yaw, roll, pitch, and shake, and set the angular velocity to 3.14 $rad/s$ and the angular velocity to 8 $rad/s^2$ to ensure that each vibration The amplitude and size remain the same and continue. The 12 objects are shaken in 4 different actions, 12 times each time. 

% We try to make the types of these objects as diverse as possible, and the objects in the container are relatively crisp, and the sound of some objects is relatively dull, as shown in Figure \ref{Waveform difference analysis}. Under different actions, we drew waveform diagrams of typical objects to show the possibility of classification, and some objects have similar shaking sounds, which brought challenges to our perception and manipulation, and made them as realistic as possible.


\subsection{Manipulation Instruction Design}

We consider the scenario where some identical bottles containing different objects are on the table. Three types of manipulation instructions are generated as shown in TABLE \ref{Problem type setting}, and are named as existence instruction, classification instruction and exploratory instruction respectively. The instructions are corresponded to different scenes. In our scene setting, one scene is associated with at least one type of instruction depending on the complexity of the scene. 

% The entire experimental scene and the setting of problem coding are mainly divided into two modes: easy and difficult. Table \ref{Problem type setting} lists the instruction model we set, where the yellow part represents the simple scene, and the green part represents the difficult scene. Images, questions, and corresponding answers are accompanied by matching semantic representations. In our scene settings, scenes and questions are not in a one-to-one correspondence, but each of our scenes is associated with at least one functional question. And point to the relevant area in the image.

A varying degree of interaction between the robot and the environment is required for different manipulation instructions. For the existence instruction, the robot only needs to recognize one target object. And for the classification instruction, the robot needs to explore all bottles until all the target objects are placed in the referred location, which is likely to involve more actions than the existence instruction. For the exploratory instruction, the robot is required to recognize the spatial relationship and manipulate with the referred object. 

% Fig. \ref{Audio-visual system verification experiment} gives an overview of the collected dataset, which includes visual information of the scene, collected sound data of each bottle, and given instructions. 

\begin{figure}[h] %%???????????????????????bp????
	\centering  %?????????
	\includegraphics[width=0.9\linewidth]{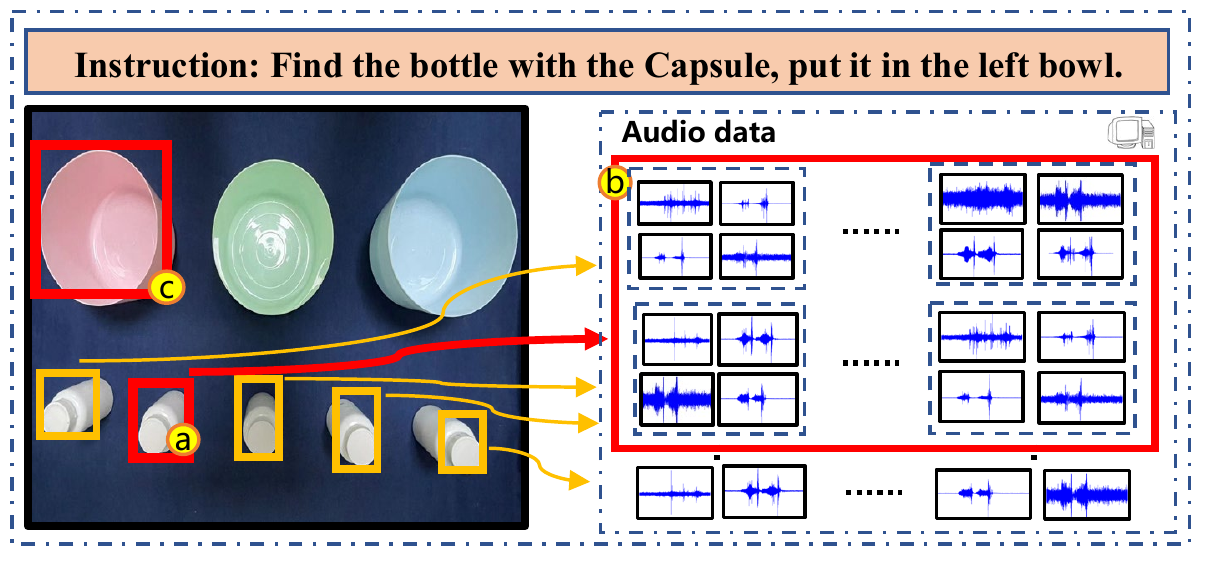} %%???????
	\caption{Audio-visual dataset. For each bottle \textbf{a} in the scene, we can pick its corresponding audio data \textbf{b} from a pool of its sound data. And the target location \textbf{c} is able to be obtained by the given instruction and the visual data of the scene.}
	\label{Audio-visual system verification experiment}
\end{figure}

\section{EXPERIMENTAL RESULTS}
\label{EXPERIMENTAL RESULTS}
We conduct both the offline and online experiments to verify the performance of the proposed task.

\subsection{Offline experiment}
To conduct the offline experiment, we utilize the collected dataset for verification. Fig. \ref{Audio-visual system verification experiment} gives an overview of the collected dataset, which includes visual information of the scene, collected sound data of each bottle, and given instructions. For each bottle \textbf{a} in the scene, we can pick its corresponding audio data \textbf{b} from a pool of its sound data. And the target location \textbf{c} is able to be obtained by the given instruction and the visual data of the scene. Therefore, we design the following offline experiments with the collected datset. 

\subsubsection{Auditory model recognition experiment}
To evaluate the recognition performance of the proposed auditory model, a sound recognition experiment is conducted. In the test phase, the bottles are filled with 12 types of objects with a random capacity between 20\% to 80\%. And then the robot manipulates with each bottle with the four predefined actions, namely yaw, roll, pitch, and shake. And the sound of the object during the manipulation is recorded. For each object, we collect the sound data for 20 times. The collected data is then fed into the auditory model for recognition. 

After evaluation, an average accuracy of 72.9\% (TABEL \ref{action_acc}) is obtained for the sound recognition task. 
%Fig. \ref{Confusion matrix} demonstrates the confusion matrix of the obtained results. 
It can be seen that the recognition accuracy of different objects varies. The cicada slough and empty bottle have a relatively low accuracy. It is because that the sound generated by them is weak and might be covered by the noise generated when the robot is running. We believe that a stronger denoising method could improve the accuracy for these two situations. 

% \begin{figure}[h] %%???????????????????????bp????
% \centering  %?????????
% \includegraphics[width=0.6\linewidth]{figure/matrix.png} %%???????
% \caption{Confusion matrix}
% \label{Confusion matrix}
% \end{figure}

% \begin{figure}[h] %%???????????????????????bp????
% 	\centering  %?????????
% 	\includegraphics[width=0.6\linewidth]{figure/jieguo.pdf} %%???????
% 	\caption{Audio classification results}
% 	\label{Audio classification results}
% \end{figure}

\subsubsection{Auditory action comparison experiment}
In the proposed framework, we use the sound generated by all the four actions to recognize the object. In this experiment, we compare the recognition results that generated by separate actions and the proposed method which utilizes a combination of four actions. For separate actions, only the sound generated by one action is used to recognize the object. The results are demonstrated in TABEL \ref{action_acc}.

% \begin{figure}[h] %%???????????????????????bp????
% 	\centering  %?????????
% 	\includegraphics[width=0.8\linewidth]{figure/accuracy.jpg} %%???????
% 	\caption{Accuracy corresponding to different actions}
% 	\label{Accuracy of different discrimination methods}
% \end{figure}

\begin{table}[]
	\caption{Accuracy corresponding to different actions}  % ????
	\label{action_acc}
	\begin{tabular}{llllll}
		\toprule  % ?????
		& all actions & \multicolumn{1}{r}{Pitch} & \multicolumn{1}{r}{Yaw} & \multicolumn{1}{r}{Roll} & \multicolumn{1}{r}{Shake} \\
		\hline
		Capsule       & \textbf{71.50}\%     & 63.60\%                   & 66.70\%                 & 60.90\%                  & 59.10\%                   \\
		Alcohol       & \textbf{62.20}\%     & 56.30\%                   & 57.90\%                 & 52.10\%                  & 49.90\%                   \\
		Red Dates     & \textbf{75.50}\%     & 62.10\%                   & 62.70\%                 & 59.30\%                  & 58.20\%                   \\
		Tablet        & \textbf{78.50}\%     & 71.50\%                   & 76.90\%                 & 72.10\%                  & 73.30\%                   \\
		Hawthorn      & \textbf{82.30}\%     & 73.10\%                   & 68.40\%                 & 70.30\%                  & 64.20\%                   \\
		Pill          & \textbf{85.40}\%     & 76.10\%                   & 81.20\%                 & 73.60\%                  & 68.50\%                   \\
		Seman Cassiae & \textbf{65.70}\%     & 59.30\%                   & 61.50\%                 & 63.60\%                  & 56.10\%                   \\
		Oyster        & \textbf{72.40}\%     & 61.80\%                   & 64.80\%                 & 59.70\%                  & 56.30\%                   \\
		Wax Pill      & \textbf{76.30}\%     & 65.80\%                   & 63.60\%                 & 65.50\%                  & 66.10\%                   \\
			Cicada Slough & \textbf{67.20}\%     & 65.10\%                   & 62.10\%                 & 58.70\%                  & 63.30\%                   \\
			Particle      & \textbf{76.40}\%     & 72.10\%                   & 74.50\%                 & 67.20\%                  & 68.40\%                   \\
			Empty         & \textbf{61.30}\%     & 53.20\%                   & 59.70\%                 & 56.70\%                  & 61.80\%    \\
		\hline
		\textbf{Average} &\textbf{72.90}\%&
		65.00\%&
		66.70\%&
		63.30\%&
		62.10\%\\
		\bottomrule

	\end{tabular}
\end{table}

\begin{figure*}[h] %%???????????????????????bp????
	\centering  %?????????
	\includegraphics[width=0.8\linewidth]{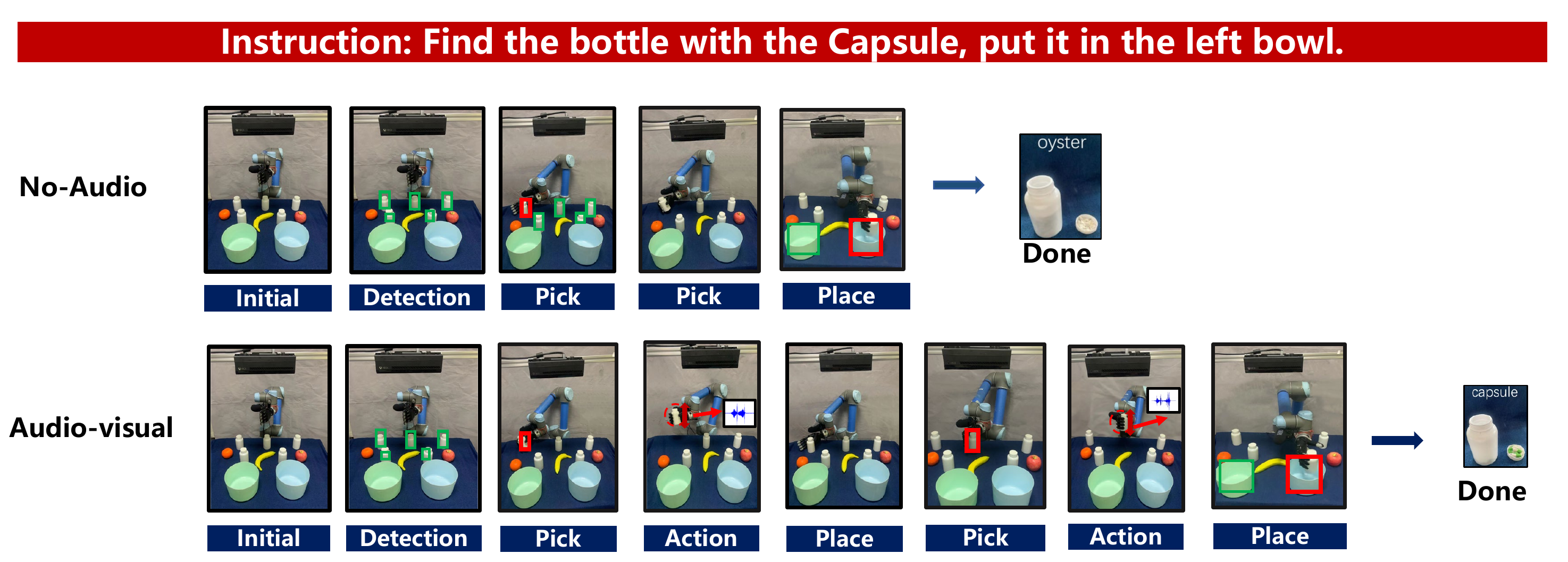} %%???????
	\caption{Online experiment comparison. The no-Audio module is a process of randomly selecting a bottle. Our model is combined with the audio-visual model, and it will not stop picking until the target object is found.}
	\label{online}
\end{figure*}

It can be seen that the results obtained from all actions is superior to that from single action. It is because that with the execution of all actions, more audio information can be collect which helps the robot identify the target object more accurately. It is noted that although the accuracy of a single action is not as good as all actions, it can still recognize some visually indistinguishable object illustrating the effectiveness of auditory information.

\subsubsection{Audio-visual system verification experiment}
\begin{figure}[h] %%???????????????????????bp????
	\centering  %?????????
	\includegraphics[width=0.7\linewidth]{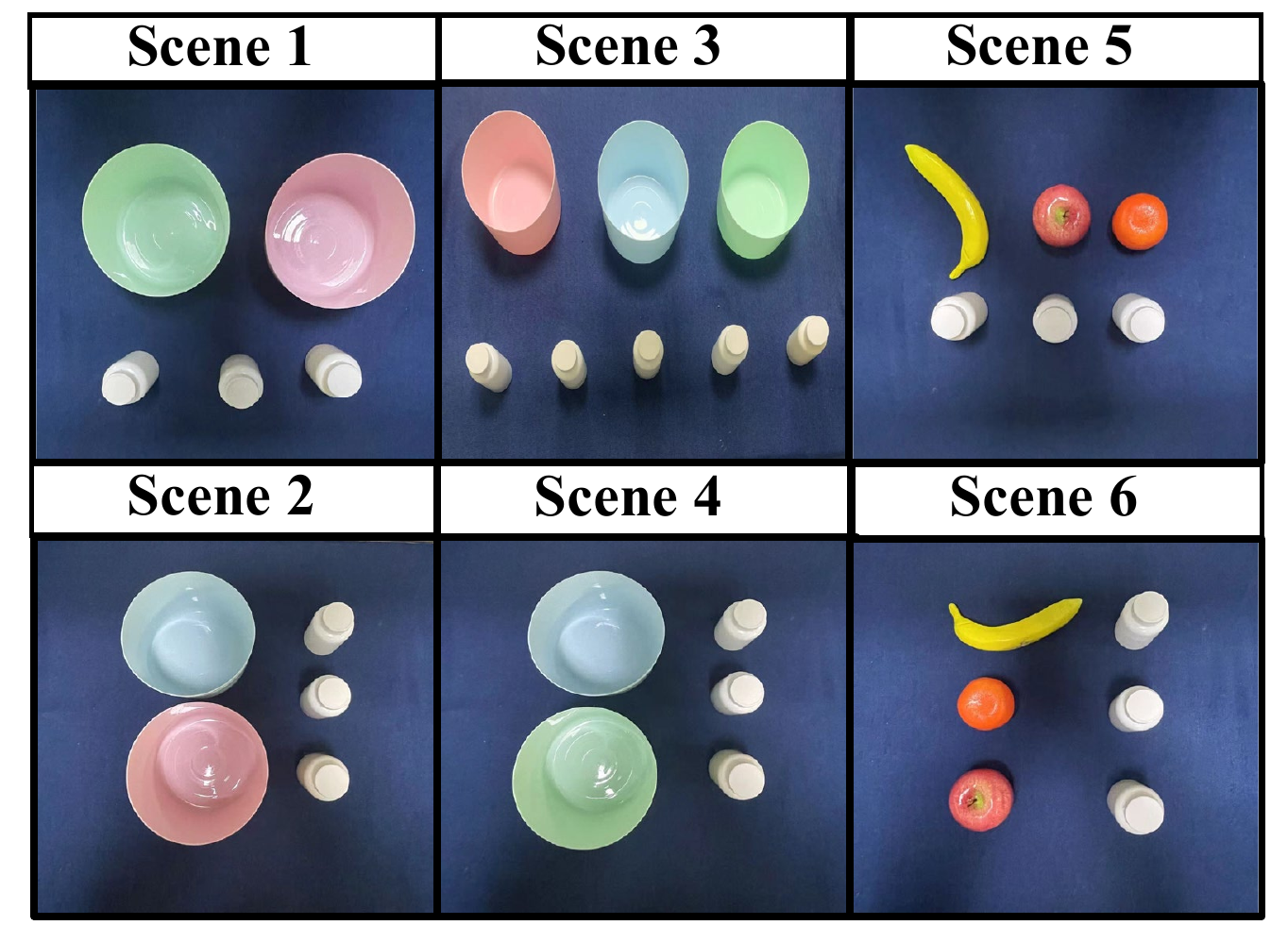} %%???????
	\caption{Example test scene. According to the position and attribute relationship between objects, it is divided into 6 types of typical scenes}
	\label{Example test scenario}
\end{figure}

To evaluate the performance of the entire audio-visual system, we have designed some test scenes as shown in Fig. \ref{Example test scenario}. Also, corresponding manipulation instructions are generated. Given the manipulation instruction, we record the scene information and the sound data generated when the robot is manipulating with the object. And three metrics are designed for a quantitative evaluation. The detailed definition is as follows. 

\begin{itemize}
	\item Target recognition accuracy (TRA): According to the visual pictures and natural language instructions in all scenes, the corresponding specific target position is generated through our instruction expression model, and the target recognition accuracy rate is calculated.
	\item Audio recognition accuracy (ARA): According to the multiple bottles that appear in the corresponding scene, we select the audio data of the bottle at the corresponding position, and calculate the audio recognition accuracy rate under different command tasks through the audio model.
	\item Overall task success rate (OTSR): This indicator is expressed as the success rate of the entire task when the first two tasks are successful.
\end{itemize}

Fig. \ref{Example test scenario} demonstrates six typical test scenes. For each scene and each instruction type, 144 manipulation instructions are given. The results of the three metrics are demonstrated in  TABLE \ref{table4}.

% For the task of the entire offline experiment, we designed the task accuracy based on three evaluation indicators, and verified it in the scenario we designed. For each experimental scenario, we tested the target accuracy and the audio recognition accuracy separately. In the case of success, the results of calculating the overall task success rate are shown in Tab. \ref{table4}:
\begin{table}[h]
	\centering  % ???????
	\caption{Audio-visual system comparison of overall performance}  % ????
	\label{table4}  % ?????????
	%??????????|?????
	% l??????c?????r?????
	\begin{tabular}{ccccc}  
		\toprule  % ?????
		Scene type&manipulation type&TRA&ARA&OTSA \\  % ????????&???\\?????
		\hline
		%&&&&\\
		scene1&Existence&87.5\%&71.4\%&68.3\% \\
		
		scene2&Existence&62.5\%&68.7\%&54.2\% \\
		
		scene3&Classification&51.7\%&42.4\%&39.6\% \\
		
		scene4&Classification&70.8\%&56.4\%&48.5\% \\
		
		scene5&Exploratory&45.8\%&64.2\%&40.1\% \\
		
		scene6&Exploratory&41.6\%&66.5\%&35.4\% \\
		
		\bottomrule
	\end{tabular}
\end{table}

For existence instruction, it only requires to recognize one target object, and thus its audio recognition accuracy is higher than that of classification instruction which requires to recognize all the target objects. For exploration instruction, it is faced with a more complex scene, and the target recognition accuracy is not as good as that of existence or classification instruction. As only one target object is required, the audio recognition accuracy is still better than the classification instruction.

\subsection{Online experiment}
We apply the proposed audio-visual framework on the real world robotic platform. To further illustrate the performance of the proposed framework, we also design a non-auditory baseline. In the non-auditory baseline, a uniform sampling method is used to select the item and complete the task. 

Fig. \ref{online} demonstrates the manipulation process when the instruction is given with both the no-audio and the proposed audio-visual approach. We also calculate the OTSR score for each method in Scene 1 and Scene 2. The audio-visual method has the OTSR of 45.4\% and 41.2\% respectively. And the no-audio method has the OTSR of 24.7\% and 22.3\% respectively. It can be seen that without the auditory model, the robot randomly selects objects, which significantly reduces the success rate of the entire task. The experimental system with auditory module can guarantee a relatively stable task success rate even in the actual environment.

% In order to prove the practicability of our audio-visual system,  we designed a non-auditory detection module under the same scene and instructions, using uniform sampling to select items to complete the entire task. Due to our purpose is to verify the entire system, and the grasping task is not the focus of our research. When our five-finger dexterous hand fails to grasp, we choose to put the target object in the palm to ensure the smooth progress of the experiment. 

% In order to facilitate the verification of the experiment, we selected two sets of different experimental scenes and the corresponding existence operation instructions.The experimental results are as follows Tab. \ref{table5} :

% \begin{table}[h]
% 	\centering  % ???????
% 	\caption{Task accuracy}  % ????
% 	\label{table5}  % ?????????
% 	%??????????|?????
% 	% l??????c?????r?????
% 	\begin{tabular}{ccc}  
% 		\toprule  % ?????
% 		&Scene1&Scene2 \\  % ????????&???\\?????
% 		\hline
% 		no-Audio& 24.7\% & 22.3\% \\
		
% 		Audio-Visual& 45.4\% & 41.2\% \\

% 		\bottomrule
% 	\end{tabular}
% \end{table}

\section{CONCLUSION}
\label{CONCLUSION}
In this paper, we develop a novel task of audio-visual grounding referring expression for robotic manipulation. Both the audio and visual information is used for understanding the referring expression in the manipulation instruction. And an audio-visual framework is proposed. We have also establish a dataset which contains auditory data, visual datat and manipulation instructions. Extensive experiments are conducted both offline and online demonstrating that with the multi-modal data, the robot performs better than with only visual data. For the future work, we would like to extend the proposed task to a more complex scenario, and propose an end-to-end framework for the robot to follow manipulation instructions.

\newpage

\appendices

\bibliographystyle{unsrt}%?????????
\bibliography{reference}  %????????Ref

\end{document}